\PassOptionsToPackage{table}{xcolor}
\documentclass[10pt,twocolumn,letterpaper]{article}

\usepackage{cvpr}              % To produce the CAMERA-READY version

\usepackage{algorithm}
\usepackage{algorithmic}
\usepackage{amsmath,amsfonts,bm}
\usepackage{color}
\usepackage{amsthm}
\usepackage{graphicx}

\usepackage{multirow}

\usepackage{makecell}
\definecolor{light-gray}{gray}{0.9}

\usepackage{xspace}

\newcommand{\tabref}[1]{\tablename~\ref{#1}}

\newcommand{\bheading}[1]{{\vspace{0\baselineskip}\noindent{\textbf{#1}}}}
\newcommand{\lheading}[1]{{\vspace{0\baselineskip}\noindent{\textit{#1}}}}

\usepackage{amsmath,amsfonts,bm}

\def\Figref#1{Figure~\ref{#1}}

\def\eqref#1{equation~\ref{#1}}
\def\Algref#1{Algorithm~\ref{#1}}

\def\1{\bm{1}}

\def\vf{{\bm{f}}}

\def\vh{{\bm{h}}}

\def\vp{{\bm{p}}}

\def\vw{{\bm{w}}}
\def\vx{{\bm{x}}}
\def\vy{{\bm{y}}}

\DeclareMathAlphabet{\mathsfit}{\encodingdefault}{\sfdefault}{m}{sl}
\SetMathAlphabet{\mathsfit}{bold}{\encodingdefault}{\sfdefault}{bx}{n}

\usepackage[accsupp]{axessibility}  % Improves PDF readability for those with disabilities.

\theoremstyle{definition}
\newtheorem{definition}{Definition}[section]

\definecolor{cvprblue}{rgb}{0.21,0.49,0.74}
\usepackage[pagebackref,breaklinks,colorlinks,allcolors=cvprblue]{hyperref}

\def\paperID{9180} % *** Enter the Paper ID here
\def\confName{CVPR}
\def\confYear{2025}

\title{MOS-Attack: A Scalable Multi-objective Adversarial Attack Framework}

\author{Ping Guo$^{1,2}$, Cheng Gong$^{1,2}$, Xi Lin$^{1,2}$, Fei Liu$^{1,2}$, Zhichao Lu$^{1,2}$, Qingfu Zhang$^{1,2}$\thanks{Corresponding author: {\tt\small qingfu.zhang@cityu.edu.hk}}, Zhenkun Wang$^3$\\
 $^1$City University of Hong Kong; \ \ $^2$CityU Shenzhen Research Institute; \\ $^3$Southern University of Science and Technology\\
}

\begin{document}
\maketitle
\begin{abstract}
    Crafting adversarial examples is crucial for evaluating and enhancing the robustness of Deep Neural Networks (DNNs), presenting a challenge equivalent to maximizing a non-differentiable 0-1 loss function.
    However, existing single objective methods, namely adversarial attacks focus on a surrogate loss function, do not fully harness the benefits of engaging multiple loss functions, as a result of insufficient understanding of their synergistic and conflicting nature.
    To overcome these limitations, we propose the Multi-Objective Set-based Attack (MOS Attack), a novel adversarial attack framework leveraging multiple loss functions and automatically uncovering their interrelations.
    The MOS Attack adopts a set-based multi-objective optimization strategy, enabling the incorporation of numerous loss functions without additional parameters.
    It also automatically mines synergistic patterns among various losses, facilitating the generation of potent adversarial attacks with fewer objectives.
    Extensive experiments have shown that our MOS Attack outperforms single-objective attacks. Furthermore, by harnessing the identified synergistic patterns, MOS Attack continues to show superior results with a reduced number of loss functions. Our code is available at \url{https://github.com/pgg3/MOS-Attack}.
\end{abstract}
\section{Introduction}

% \fei{ just some thoughts, Title: MOS-Attack: A Scalable Multi-objective Adversarial Attack Framework with Automatic Multi-loss Analysis}

% \xl{Or might be just: MOS-Attack: A Scalable Multi-objective Adversarial Attack Framework}

% \luz{Title is too long. ``A Scalable Multiobjective Adversarial Attack Framework with Pareto Set Learning''}

Deep neural network (DNN) models have significantly advanced the field of computer vision~\cite{he:2016:deep,sabour:2017:dynamic,deng:2009:imagenet,CIFAR10},
yet they are vulnerable to adversarial examples~\citep{szegedy:2014:intriguing,goodfellow:2015:explaining}.
Such examples are inputs that have been subtly modified to cause misclassification, potentially leading to catastrophic consequences in real-world scenarios~\cite{deng:2019:efficient,cao:2019:adversarial,laleh:2022:adversarial}.
Consequently, the development of sophisticated adversarial attack algorithms is crucial for evaluating and enhancing the robustness of these models~\cite{croce:2020:reliable, madry:2018:towards, zhang:2019:theoretically}.
However, devising these algorithms presents inherent challenges due to the non-differentiable nature of the original optimization problem,
necessitating the use of surrogate loss functions~\cite{gowal:2019:alternative} to facilitate gradient-based adversarial attacks~\cite{goodfellow:2015:explaining,madry:2018:towards,croce:2020:reliable}.

The metric for measuring misclassification is the non-differentiable 0-1 loss function, which surrogate loss functions endeavor to approximate~\cite{li:2022:autoloss}.
Adversarial attacks are designed to generate a perturbation $\bm{\delta}$ that causes the misclassification of an input $\vx$ with its corresponding label $\vy$.
This can be formulated as~\cite{madry:2018:towards,gowal:2019:alternative}:
\begin{equation}\label{eq:org_problem}
    \max_{\bm{\delta}\in\mathcal{B}} L_{\text{0-1}}(\vh_\theta(\vx+\bm{\delta}), \vy),
\end{equation}
where $\vh_{\theta}$ represents the DNN model parameterized by $\theta$, $L_{\text{0-1}}$ denotes the 0-1 loss function, and $\mathcal{B}$ is the set of allowable perturbations.

Considering the computational intractability of the problem in Equation~(\ref{eq:org_problem})~\cite{arora:1997:hardness},
contemporary research commonly employs a differentiable surrogate loss function in place of the 0-1 loss function.
This approach enables the utilization of gradient-based optimization techniques to address the resultant surrogate optimization problem.
It has propelled considerable progress in gradient-based algorithms, including the Fast Gradient Sign Method (FGSM)~\cite{goodfellow:2015:explaining}, Projected Gradient Descent (PGD)~\cite{madry:2018:towards}, and Carlini \& Wagner (C\&W) attack~\cite{carlini:2017:towards}.
Notably, the versatility of the PGD attack has increased with the adoption of diverse surrogate loss functions (APGD-CE, APGD-DLR)~\cite{sriramanan:2020:guided}
and the integration of sophisticated optimization techniques (ACG)~\cite{yamamura:2022:diversified}.
These developments have given rise to more sophisticated adversarial methods.

While single-objective attacks have attracted considerable attention,
there is an emerging trend towards integrating multiple loss functions to bolster the attack's efficacy.
Some early endeavors include using multiple targeted loss functions to guide untargeted attack~\cite{gowal:2019:alternative} and the strategic alternation of loss functions in the attack process\cite{nikolaos:2022:alternating}.
Furthermore, the adoption of diverse surrogate loss functions such as GAMA~\cite{sriramanan:2020:guided}, BCE~\cite{wang:2020:improving}, and DLR~\cite{croce:2020:reliable} has been instrumental in advancing adversarial attacks.

Despite the potential advantages for incorporating multiple loss functions, direct optimization with a vast array of adversarial examples is inefficient.
Moreover, the methodology for targeting suitable loss functions to mount effective adversarial attacks is lacking.
Therefore, it is imperative to develop a scalable framework that can efficiently coordinate multiple surrogate loss functions, concentrate on a limited subset, and reduce the number of adversarial examples needed for optimization.

% the field still lacks a unified and scalable framework to coordinate multiple surrogate loss functions and systematically explore their synergistic and conflicting interactions.

% \xl{A stronger argument might be needed to clearly point out the shortcomings of the current methods, and strongly motivate the need for our method (e.g.,  why we need "a unified and scalable framework to coordinate multiple surrogate loss functions and systematically explore their synergistic and conflicting interactions").}

To bridge this knowledge gap, we introduce the Multi-Objective Set-based Attack (MOS Attack), a novel framework for conducting multi-objective adversarial attacks and investigating the interactions among various surrogate loss functions.
Our framework notably offers: \textit{1)} a scalable, parameter-free template for executing multi-objective adversarial attacks, and \textit{2)} automated method for the discovery of synergistic loss patterns.
The MOS Attack employs a suite of surrogate loss functions and initiates an adaptable number of adversarial examples, thereby defining a smooth set-based optimization problem.
Subsequently, single-objective gradient-based optimization techniques, which require only minimal adjustments, can efficiently address this problem.
Following the optimization phase, an automated analysis identifies synergistic patterns within the adversarial examples.
These patterns enable the construction of powerful multi-objective adversarial attacks that require fewer objectives, allowing a more efficient allocation of computational resources to each objective.

We have implemented our approach using four widely recognized surrogate loss functions as outlined in previous research~\cite{sriramanan:2020:guided,wang:2020:improving}, as well as four additional functions identified through extensive loss function searches~\cite{li:2022:autoloss,xia:2024:tightening}.
The resulting MOS-8\footnote{We use MOS-8 Attack to denote MOS Attack implemented with 8 loss functions, MOS-3$^*$ to denote the attack implemented with three selected loss functions.} Attack has proven highly effective through extensive experimentation on the CIFAR-10~\cite{CIFAR10} and ImageNet~\cite{deng:2009:imagenet} datasets, outperforming state-of-the-art methods that leverage advanced gradient-based optimization or eight distinct single-objective attacks for each surrogate loss.
Moreover, by examining the synergistic patterns uncovered by MOS-8 Attack, we have developed a tri-objective attack, MOS-3$^*$, which has also shown superior performance.
%  \fei{confused on MOS-8 and MOS-3$^*$}

Our contributions can be summarized as follows:
\begin{itemize}
    \item We introduce the first multi-objective adversarial attack framework, the MOS Attack, which tackles the challenge of generating adversarial examples with multiple loss functions. This framework is parameter-free and readily extensible with new loss functions.

    \item Our framework also offers an automated method for identifying synergistic patterns among loss functions, which can be used to construct powerful multi-objective attacks with fewer objectives, facilitating a more efficient allocation of computational resources.

    \item We have implemented our framework with 8 loss functions to form the MOS-8 Attack, which has been extensively tested on CIFAR-10 and ImageNet datasets. Additionally, synergistic analysis over these 8 loss functions has been conducted to provide insights regarding their interactions and led to the development of a powerful tri-objective attack, MOS-3$^*$.
\end{itemize}

\section{Background}

%\subsection{Adversarial Attack}
% \fei{should we mention black and white box attacks, are MOS applicable to both settings}

Adversarial attacks encompass methods that create adversarial examples,
which are used to assess and enhance model robustness~\cite{madry:2018:towards,croce:2020:reliable}.
A white-box threat model is often considered for evaluating adversarial robustness, where the adversary has full access to the model's architecture, parameters, and gradients.
While white-box existing strategies mainly focus on one surrogate loss function~\cite{antoniou:2020:square,guo:2024:lautoda,fu:2022:autoda,yamamura:2022:diversified}, a recent trend is the integration of multiple loss functions into the attack paradigm~\cite{wang:2024:amulti,deist:2023:multi,Suzuki:2019:adversarial,baia:2021:effective,liu:2024:effective}.

% \xl{The term Single-*o*bjective and Multi-*O*bjective should be consistent in this paper.}

\bheading{Single-Objective Attacks.}
White-box attack methodologies typically employ a singular surrogate loss function, focusing on optimization to craft adversarial examples.
Established strategies include the FGSM~\cite{goodfellow:2015:explaining}, C\&W attack~\cite{carlini:2017:towards}, and PGD attack~\cite{madry:2018:towards}.
Croce~\etal proposed a novel parameter-free approach, Auto-PGD (APGD) attack, utilizing both Cross Entropy (CE) and the Difference of Logits Ratio (DLR) loss functions.
These were subsequently incorporated into the AutoAttack framework as APGD-CE and APGD-DLR~\cite{croce:2020:reliable}.
Expanding upon this, Yamamura~\etal enhanced APGD with conjugate gradient techniques, resulting in the creation of the powerful Auto Conjugate Gradient (ACG) attack~\cite{yamamura:2022:diversified}.

\begin{figure*}[t]
    \centering
    \begin{subfigure}{0.32\linewidth}
        \includegraphics[width=0.9\textwidth]{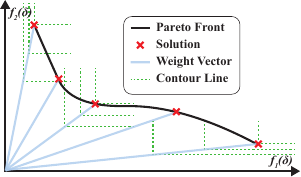}
        \caption{Decomposition-based Optimization.}
        \label{fig:opt-a}
    \end{subfigure}
    \hfill
    \begin{subfigure}{0.32\linewidth}
        \includegraphics[width=0.9\textwidth]{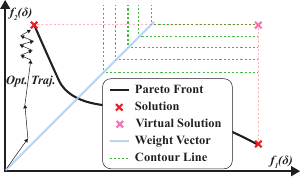}
        \caption{Set-based Optimization.}
        \label{fig:opt-b}
    \end{subfigure}
    \begin{subfigure}{0.32\linewidth}
        \includegraphics[width=0.9\textwidth]{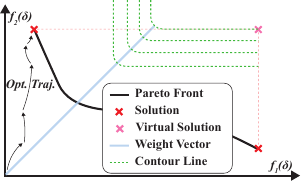}
        \caption{Smooth Set-based Optimization (MOS).}
        \label{fig:opt-c}
    \end{subfigure}
    \vspace{-0.75\baselineskip}
    \caption{Comparison of different optimization methods for conducting multi-objective adversarial attacks.}
    \vspace{-0.75\baselineskip}
\end{figure*}

\bheading{Multi-Objective Attacks.}
Recent advancements in adversarial research have involved the integration of multiple surrogate loss functions into the attack framework.
Gowal~\etal introduced multiple targeted losses to enhance untargeted PGD attacks~\cite{gowal:2019:alternative}.
Further, work by Nikolaos~\etal established that the strategic variation of surrogate loss functions considerably improve adversarial attack performance~\cite{nikolaos:2022:alternating}.
However, these studies typically lack a systematic approach and a solid theoretical underpinning for managing multiple losses.

Concurrently, researchers have expanded the adversarial attack framework by introducing other types of objectives.
Williams~\etal investigated the inclusion of additional norm constraints~\cite{williams:2023:black-box}, while Guo~\etal and Liu~\etal have investigated the trade-off between perturbation intensity and confidence measures~\cite{guo:2024:exploring,liu:2024:effective}.
These efforts have contributed to the development of more diversified attack methodologies.

Our approach represents the first attempt to systematically incorporate multiple loss functions into adversarial attacks and optimize the corresponding multi-objective optimization problem using a minimal set of examples via smooth set-based optimization techniques.

\section{Multi-Objective Set-based Attack}
In this section, we propose the problem formulation of the smooth set-based approach for multi-objective adversarial attacks.
% \xl{Can we use more active terms, such as "In this section, we propose..."?}
We begin by defining the multi-objective adversarial attack, which employs multiple surrogate loss functions, as a multi-objective optimization problem.
Subsequently, we introduce the decomposed subproblems and identify three optimization challenges.
Finally, we propose the formulation of the smooth set-based optimization problem as a solution to the challenges posed by the multi-objective nature of adversarial attacks.

\subsection{Multi-Objective Adversarial Attack}
This study seeks to simultaneously optimize multiple surrogate loss functions, rather than relying on a singular loss function, to craft adversarial examples.
Given $m$ loss functions $L_1, \ldots, L_m$
% \fei{$\ell_i, \ell_{01}, \ell_\infty$ might misleading}
, we define a multi-objective optimization problem as follows:
\begin{equation}
    \begin{aligned}
         & \max_{\bm{\delta}\in \mathcal{B}} \vf(\bm{\delta}) = (f_1(\bm{\delta}),\ldots, f_m(\bm{\delta})),       \\
         & f_i(\bm{\delta})                   =  L_i(\vh_\theta(\vx+\bm{\delta}), y), \forall i\in\{1,\ldots, m\}.
    \end{aligned}
\end{equation}
The notation remains consistent with the single-objective scenario as depicted in Equation~(\ref{eq:org_problem}). Furthermore, this paper adopts the extensively utilized $\ell_\infty$-ball as the constraint set for perturbations, denoted by $\mathcal{B} = \{\bm{\delta}: \|\bm{\delta}\|_\infty \leq \epsilon\}$.

Examples with higher values across multiple surrogate loss functions are more susceptible to misclassification by the model.
In existing literature, this statement is supported by frequent misclassifications of adversarial examples with high values on singular loss functions~\cite{madry:2018:towards,croce:2020:reliable,yamamura:2022:diversified}.

Nevertheless, since there is often no single solution that maximizes all the loss functions simultaneously, a set of best trade-off solutions becomes necessary.
This set of solutions is called the Pareto set, and the corresponding values of the objective functions are called the Pareto front.
A formal description of Pareto optimality is delineated below:

\begin{definition}[Pareto Optimal]
    A solution $\bm{\delta}^*$ is Pareto optimal if there is no other solution $\bm{\delta}$ such that $f_i(\bm{\delta}) \geq f_i(\bm{\delta}^*)$ for all $i\in\{1,\ldots, m\}$ and $f_i(\bm{\delta}) > f_i(\bm{\delta}^*)$ for at least one $i\in\{1,\ldots, m\}$.
\end{definition}

\begin{definition}[Pareto Set and Pareto Front]
    The Pareto set is the set of all Pareto optimal solutions, and the Pareto front is the set of all the values of the objective functions at the Pareto optimal solutions.
\end{definition}

\subsection{Decomposition-Based Optimization}
In this research, we employ the Tchebycheff decomposition to transform the multi-objective problem into a suite of single-objective subproblems.
Contrary to the linear scalarization method, the Tchebycheff approach is capable of targeting any location on the Pareto front.
This is well-recognized in the discipline of multi-objective optimization~\cite{deb:2002:fast,zhang:2007:moead,miettinen2012nonlinear}.
% \xl{For readers not familiar with MOO, a common question could be "why not simple linear scalarization?".}
By adopting this method, given $K$ weight vectors $\vw_1, \ldots, \vw_K$, the $k$-th decomposed subproblem is defined as follows:
\begin{equation}
    \begin{aligned}\label{eq:decomp_problem}
         & \max_{\bm{\delta}\in \mathcal{B}} g_k(\bm{\delta}|\vw_k)  = \min_i w_{ki} |f_i(\bm{\delta}) -z^*_i|,
    \end{aligned}
\end{equation}
where $w_{ki}$ is the $i$-th element of the $k$-th weight vector, and $z^*_i$ denotes the ideal value for the $i$-th objective. Upon solving these subproblems, a set of solutions correlated with the weight vectors is obtained. This set can approximate the Pareto set, as illustrated in Figure \ref{fig:opt-a}.

The Tchebycheff method is instrumental in identifying both convex and non-convex parts of the Pareto front with vertical contour lines~\cite{wang:2015:pareto}, as demonstrated in Figure \ref{fig:opt-a}.
Nonetheless, it presents three challenges in optimization:
\begin{itemize}
    \item \textbf{Complexity:} Accurate approximation of the Pareto front necessitates multiple points, exceeding the number of objectives ($>m$).
    \item \textbf{Ambiguity:} The selection of appropriate weight vectors is challenging.
    \item \textbf{Non-differentiability:} The function $g_k$ contains non-differentiable points.
\end{itemize}

% We use smooth set-based optimization to circumvent these challenges.

\subsection{Smooth Set-based Optimization} \label{subsec:smoothset}
To address the challenges associated with Tchebycheff decomposition, we propose a formulation that leverages a smooth set-based optimization approach.
The primary issue is the number of adversarial examples needed to maximize the surrogate loss functions.
We tackle this by deliberately selecting a set of $ K $ adversarial examples, with $ K < m $.
Additionally, we default the weight vector to an all-ones configuration to eliminate the ambiguity in selecting the weight vector.
Lastly, we smooth the optimization problem to circumvent non-differentiability issues.

\bheading{Set-based Optimization}
Suppose we have a set of $K$ adversarial examples $\bm{\Delta}=\{\bm{\delta}_1,\ldots,\bm{\delta}_K\}$ to accommodate multiple objectives and one weight vector $\vw$ for specifying the contour lines.
The set-based optimization problem can be formulated as:
\begin{equation}
    \begin{aligned}\label{eq:set_based_decomp}
         & \max_{\bm{\Delta}} g(\bm{\Delta}|\vw)  = \min_i w_i |\max_{k_i} f_i(\bm{\delta}_{k_i}) -z^*_i|,
    \end{aligned}
\end{equation}
% \xl{Why "s.t."? Why not just simply $\max_{\bm{\Delta}} \min_i w_i |\max_{k_i} f_i(\bm{\delta}_{k_i}) -z^*_i|$ (with proper modification to make $\bm{\Delta}=\{\bm{\delta}_1,\ldots,\bm{\delta}_K\}$ more clear)?}
where $w_i$ is the $i$-th element of the weight vector, and $z^*_i$ is the optimal value of the $i$-th objective function.

\lheading{A Geometric Interpretation.}
The inner maximization problem as formulated in Equation~(\ref{eq:set_based_decomp}) allows each perturbation vector, $\bm{\delta}\in\bm{\Delta}$, to impart its dimensionality upon the objective function.
We conceive a 'virtual adversarial example' as a combination of the most advantageous dimensional attributes of adversarial examples.
The essence of the set-based optimization procedure lies in pushing this virtual adversarial example towards extreme points along the contour lines, as depicted in \Figref{fig:opt-b}.

We investigate the relationship between the number of adversarial examples $K$ and the number of loss functions $m$. Specifically:
\begin{itemize}
    \item $K < m$: A smaller number of adversarial examples are utilized to address a multitude of objectives. This approach enables optimization of the functions using reduced resources. In the extreme scenario where $K=1$, a single solution must fulfill all objectives.

    \item $K = m$: there exists a theoretical optimal solution comprising the individual optimal adversarial examples for each objective function. Through proper optimization, this ideal state may be achieved.
\end{itemize}

So far, the first two challenges of decomposition-based optimization have been addressed by the set-based optimization problem.
However, the third challenge remains unresolved.
This can lead to oscillation in the optimization process, as illustrated in \Figref{fig:opt-b}.
Therefore, we need to design a smooth approximation of the set-based optimization problem.
% \xl{Indeed, the set-based optimization is not very well motivated, especially for readers not familiar with multi-objective optimization. More discussion might be needed before introducing the set-based optimization approach.}

% Although the above problem allows us to flexibly adjust the number of solutions to accommodate multiple objectives and address the first two challenges posed by the decomposition-based optimization,
% the usage of maximum operation and the minimum in the objective function makes the optimization problem not smooth and hard to optimize.
% This can lead to oscillation in the optimization process, as illustrated in \Figref{fig:opt-b}. Therefore, we need to design a smooth approximation of the set-based optimization problem.

\bheading{Smooth Set-based Optimization}
To smooth the above optimization problem, we need to take advantage of smooth max and smooth min operators~\cite{beck:2012:smoothing,lin:2024:smooth,lin:2024:few}.

\begin{equation}\label{eq:smooth_ops}
    \begin{aligned}
         & \max\left\{x_1,\ldots,x_m\right\} \approx \mu\log\left(\sum_{i=1}^m e^{x_i/\mu}\right),   \\
         & \min\left\{x_1,\ldots,x_m\right\} \approx -\mu\log\left(\sum_{i=1}^m e^{-x_i/\mu}\right),
    \end{aligned}
\end{equation}
where $\mu$ is a smoothing parameter. A proof of the above approximation can be found in~\cite{beck:2012:smoothing}.

Using the above operators, the objective function in Equation~(\ref{eq:set_based_decomp}) can be approximated as:
\begin{equation}
    \begin{aligned}
         & g(\bm{\Delta}|\vw) = \min_i w_i |\max_{k_i} f_j(\bm{\delta}_{k_i}) - z^*_i|,                                               \\
        %  & \approx \min_i w_i|\mu\log\left(\sum_{k=1}^K e^{f_i(\delta_k)/\mu}\right) -z^*_i|,                                    \\
         & \approx -\mu\log\left(\sum_i^m e^{-(w_i|\mu\log\left(\sum_{k=1}^K e^{f_i(\bm{\delta}_k)/\mu}\right) -z^*_i|)/\mu}\right) . \\
        \label{eq:smoothed_set_based_opt}
    \end{aligned}
\end{equation}

Furthermore, if we consider $z^*_i=0$ and a uniform weight vector with $w_i=w_j,\forall i,j$, we can get our final optimization problem as:
\begin{equation}
    \begin{aligned}
         & \max_{\mathbf{\Delta}}g(\mathbf{\Delta})  = -\mu\log\left(\sum_i^m (\sum_{k=1}^K e^{f_i(\bm{\delta}_k)/\mu})^{-1}\right). \\
        \label{eq:smoothed_set_based_opt_simplified}
    \end{aligned}
\end{equation}
% \xl{Why not just $\max_{\mathbf{\Delta}} -\mu\log\left(\sum_i^m (\sum_{k=1}^K e^{f_i(\bm{\delta}_k)/\mu})^{-1}\right)$?}

The above formulation eases the optimization process and avoids oscillations in the optimization process, which is analyzed in the multi-objective literature~\cite{lin:2024:smooth}.
We illustrate the smoothed set-based optimization problem along with a possible optimization trajectory in \Figref{fig:opt-c}.

% \xl{The challenge of smoothness is also not well-motivated... Since the focus of this work is on adversarial attack, I suggest to give more room to a detailed discussion to motivate the set-based adversarial attack, and left the discussion of smooth optimization in the appendix.}

\section{Methodology}
% In this section, we present the framework of MOS Attack, which contains three main steps: \textit{1)} smoothing the set-based optimization problem, \textit{2)} optimizing through gradient-based algorithms, and \textit{3)}analysing the coupling effects between the solutions.  Specifically, we first transform the set-based problem into a smoothed single-objective problem using smoothed minimum and maximum operator. Then, we implement the gradient-based method in the form of a variant of APGD algorithm. Finally, statistical features are extracted from the obtained solutions to analyze the coupling effects.
\subsection{MOS Attack: Implementation by APGD}
By formulating a smooth set-based optimization problem,
we can now apply gradient-based optimization algorithms for its efficient resolution.
Within the domain of adversarial attacks, our framework incorporates the well-known APGD algorithm~\cite{croce:2020:reliable}.
In this section, we provide a detailed explanation of our attack, as outlined in \Algref{alg:gradient}.

\begin{figure*}[t]
    \begin{minipage}[t]{0.6\textwidth}
        \captionof{table}{The loss function utilized for implementing our attack. }
        \vspace{-0.75\baselineskip}
        % \luz{(1) $\vh_y$ and $\vh_j$ are not defined. (2) Some of the losses seem incorrect: e.g., TRADES [55] add a KL divergence between the predicted logits on clean and adv examples to the CE; MART [49] also has the KL divergence component in its loss as done in [55]. (3) Missing loss: adv logit pairing loss is another widely used loss for adv attack (800 citations from I. Goodfellow \url{https://arxiv.org/pdf/1901.08573})}
        \label{tab:loss_funcs}
        \resizebox{!}{3.5\baselineskip}{
            \begin{tabular}{llc}
                \toprule
                \textbf{ID} & \textbf{Loss Function}                                                                                                                        & \textbf{Formula}                                                                             \\
                \midrule
                0           & Cross Entropy Loss~\cite{szegedy:2014:intriguing,madry:2018:towards,goodfellow:2015:explaining,kurakin:2017:adversarial,carlini:2017:towards} & $-\vh_y(\vx)+\log(\sum_{i=1}^K e^{\vh_i(\vx)})$                                              \\
                1           & Marginal Loss~\cite{croce:2020:minimally,carlini:2017:towards,gowal:2019:alternative,croce:2020:reliable,sriramanan:2020:guided}              & $-\vh_y(\vx)+\max_{j\neq y}\vh_j(\vx)$                                                       \\
                2           & Difference of Logits Ratio~\cite{croce:2020:minimally}                                                                                        & $(-\vh_y(\vx)+\max_{j\neq y}\vh_j(\vx))/(\vh_{\pi_1}(\vx)-\vh_{\pi_3}(\vx))$                 \\
                3           & Boosted Cross-Entropy Loss~\cite{wang:2020:improving}                                                                                         & $-\log\vp_y(\vx)-\log(1-\max_{j\neq y}\vp_j(\vx))$                                           \\
                4           & Searched Loss 1~\cite{xia:2024:tightening}                                                                                                    & $\sum_i\exp(10\vp/\max_j\vp_j)$                                                              \\
                5           & Searched Loss 2~\cite{xia:2024:tightening}                                                                                                    & $\exp(-\max (\text{softmax}(\vh+2\text{softmax}(5\vh))))$                                    \\
                6           & Searched Loss 3~\cite{xia:2024:tightening}                                                                                                    & $\text{softmax}(-\text{softmax}(2\exp(\vh)\vh))(\text{softmax}(2\vh)+2\vy_{\text{one-hot}})$ \\
                7           & Searched Loss 4~\cite{xia:2024:tightening}                                                                                                    & $(\text{softmax}(\text{softmax}(2\vh)+\vh-\vy_{\text{one-hot}})-\vy_{\text{one-hot}})^2$     \\
                \bottomrule
            \end{tabular}
        }
    \end{minipage}
    \begin{minipage}[t]{0.38\textwidth}
        \captionof{table}{The notations used in the loss functions. \label{tab:loss_notations}}
        \vspace{-0.75\baselineskip}
        \resizebox{!}{3.5\baselineskip}{
            \begin{tabular}{lc}
                \toprule
                \textbf{Notation}      & \textbf{Description}                                \\
                \midrule
                $\vx$                  & The adversarial example.                            \\
                $\vh$                  & The vector of logits.                               \\
                $\vh_y$                & The logit corresponding to the true class.          \\
                $\vh_j$                & The logit corresponding to the $j$-th class.        \\
                $\vh_{\pi_i}$          & The $i$-th highest logit.                           \\
                $\vp_y$                & The probability corresponding to the true class.    \\
                $\vp_j$                & The probability corresponding to the $j$-th class.  \\
                $\vy_{\text{one-hot}}$ & The one-hot vector corresponding to the true class. \\
                \bottomrule
            \end{tabular}
        }
    \end{minipage}
    % \vspace{-0.5\baselineskip}
\end{figure*}

\begin{algorithm}[t]
    \caption{MOS Attack} \label{alg:gradient}
    \begin{algorithmic}[1]
        \STATE \textbf{Input:} $g$, $\mathcal{B}$, $\mathbf{\Delta}^{(0)}$, $\eta$, $N_\text{iter}$, $W=\left\{w_0,\ldots, w_n\right\}$
        % Initial $\mathbf{X}_K^0$, Iterations $N_{iter}$, Step size $\eta_t$, Checkpoint $T=\left\{t_0, \ldots, t_n\right\}$
        \STATE \textbf{Output:} $\bm{\Delta}_\text{adv}$
        \STATE $\bm{X}^{(0)} \leftarrow \vx + \mathbf{\Delta}^{(0)}$
        \STATE $\bm{X}^{(1)} \leftarrow P_\mathcal{B}(\bm{X}^{(0)} + \eta\nabla g(\mathbf{\Delta}^{(0)}))$
        \STATE $\mathbf{\Delta}^{(1)} \leftarrow \bm{X}^{(1)} - \vx$
        \STATE $g_\text{max} \leftarrow \max \left\{g(\mathbf{\Delta}^{(0)}), g(\mathbf{\Delta}^{(1)})\right\}$
        \STATE $\bm{X}_\text{max} \leftarrow \bm{X}^{(0)}$ \textbf{if} $g_\text{max} \equiv g(\mathbf{\Delta}^{(0)})$ \textbf{else} $\bm{X}_\text{max} \leftarrow \bm{X}^{(1)}$
        \FOR{$k=1$ to $N_\text{iter}-1$}
        \STATE $\bm{Z}^{(k+1)} \leftarrow P_\mathcal{B}(\bm{X}^{(k)} + \eta\nabla g(\mathbf{\Delta}^{(k)}))$ \label{alg:line:9}
        \STATE $\bm{X}^{(k+1)} \leftarrow P_\mathcal{B}(\bm{X}^{(k)} + \alpha\left(\bm{Z}^{(k+1)} - \bm{X}^{(k)}\right) + (1-\alpha)\left(\bm{X}^{(k)} - \bm{X}^{(k+1)}\right))$ \label{alg:line:10}
        \STATE $\mathbf{\Delta}^{(k+1)} \leftarrow \bm{X}^{(k+1)} - \vx$
        \IF{$g(\mathbf{\Delta}^{(k+1)}) > g_\text{max}$}
        \STATE $\bm{X}_\text{max} \leftarrow \bm{X}^{(k+1)}$ and $g_\text{max} \leftarrow  g(\mathbf{\Delta}^{(k+1)})$
        \ENDIF
        \IF{$k\in W$}
        \IF{Condition 1 \textbf{or} Condition 2}
        \STATE $\eta \leftarrow \eta/2$ and $\bm{X}^{(k+1)} \leftarrow \bm{X}_\text{max}$ and $\mathbf{\Delta}^{(k+1)} \leftarrow \bm{X}_\text{max} - \vx$
        \ENDIF
        \ENDIF
        \ENDFOR
    \end{algorithmic}
\end{algorithm}

\bheading{Initialization.}
The initialization process involves specifying the input parameters and determining the initial adversarial examples.
We follow a procedure similar to that used in the APGD.
However, our approach take as input \text{1)} a set of adversarial examples $\bm{\Delta}$, and \text{2)} an objective function $g(\bm{\Delta})$ defined in Equation~(\ref{eq:smoothed_set_based_opt_simplified}).

\bheading{Momentum-based Update Rule.}
We adopt the same momentum-based update rule as in APGD, which is considered to be stable and efficient.
The details are delineated in lines~\ref{alg:line:9} and \ref{alg:line:10} of \Algref{alg:gradient}.
Our modification addresses the optimization of a set of adversarial examples rather than a single example.
Therefore, we have adjusted the update rule to: \text{1)} optimize $\bm{X}$ and $\bm{\Delta}$ concurrently, and \text{2)} implement a set-based projection operator.

\lheading{Optimzation Representation.}
Considering our function $g$ incorporates  $\bm{\Delta}$ and subsequent projection requires $\bm{X}$'s range, concurrent optimization of both is essential.
Notably, the statement in line~\ref{alg:line:9} consistently applies because $\nabla_{\bm{X}}g(\bm{\Delta})=\nabla_{\bm{\Delta}}g(\bm{\Delta})$, with $\bm{X} = \bm{\Delta} + \vx$.

\lheading{Set-based Projection.}
A pivotal component in gradient-based adversarial methodologies is the projection operator, which constrains the adversarial examples within the defined perturbation bounds.
In our context, the challenge entails projecting an ensemble of adversarial examples.
This is executed by individually projecting each example within the allowable perturbation boundary.

\bheading{Step Size Adjustment.}
We use the same step size control method in APGD. The initial step size $\eta$ is set to $2\epsilon$, where $\epsilon$ is the perturbation budget. When the checkpoint is reached, the following two conditions are checked:
\begin{enumerate}
    \item $N_\text{inc} < \rho (w_j - w_{j-1})$,
    \item $\eta^{w_{j-1}} = \eta^{w_j}$ and $g^{w_{j-1}}_\text{max} = g^{w_{j}}_\text{max}$,
\end{enumerate}
where $N_\text{inc}=\#\{i=w_{j-1},\ldots,w_{j}-1|g(\bm{\Delta}^{(i+1)}) > g(\bm{\Delta}^{(i)})\}$ and $g^{k}_\text{max}=\max\{g(\bm{\Delta}^{(i)})|i=1,\ldots,k\}$.

\subsection{Automated Synergistic Pattern Mining}
Few solutions automatically maximize different loss functions in groups in smooth set-based optimization~\cite{lin:2024:few}.
To mine these loss synergistic patterns, we propose an automated mining method.
This method includes two steps: \textit{1)} determining the dominant examples that contribute to the loss maximization, and \textit{2)} determining the synergistic pattern of these dominant examples.

\bheading{Determining Dominant Examples.}
With a set of $K$ perturbations $\bm{\Delta}=\left\{\bm{\delta}_1, \ldots, \bm{\delta}_K\right\}$ from the MOS Attack, we aim to identify the dominant perturbations that maximize the loss functions.
Formally, we want to find an index vector $\bm{\beta} = \left[\beta_1, \ldots, \beta_K\right], \beta_i\in\{0,1\},\forall i,$ for specifying a subset of perturbations $\bm{\Delta}_{\bm{\beta}} = \left\{\bm{\delta}_i | \beta_i = 1\right\}$ that still maximize the loss functions.

We first perform min-max normalization on the loss functions $f_i(\bm{\delta}_k), \forall i$, and then the above formulation can be rewritten as a bi-objective optimization problem:
\begin{equation} \label{eq:determine}
    \min_{\bm{\beta}} (\sum_{i=1}^m \max_{k=1}^K \bar{f}_i(\delta_k) - \max_{k=1}^K \beta_k \bar{f}_i(\delta_k), \|\bm{\beta}\|_0),
\end{equation}
where $\bar{f}_i(\delta_k)$ is the normalized loss function.
The first term serves to minimize the optimization gap. The $\ell_0$ norm, which is the number of non-zero elements in a vector, aims to minimize the number of selected examples.

\begin{table*}[t]
    \centering
    \caption{\textbf{Overall Results.} A comparative analysis of attack success rate among MOS-8 attacks with APGD-CE, ACG-CW, and APGD-All. For MOS-8 Attack, we record its $K$ value, while for others it denoted the number of restarts. Notably, for APGD-All, we have documented the index of the surrogate loss functions corresponding to the highest attack success rate. The optimal outcome is highlighted in bold and marked with a grey background. The second-best performance is underscored for emphasis. }
    % \luz{AFAIK, many existing attack methods (TRADES, MART, AutoAttack), including many of those you cited (Gowal et al., Rebuffi et al.), compare robust accuracy instead of attack success rate. Unless there is a strong motivation to compare attack success rate, I suggest following the mainstream setting to compare the robust accuracy. Even though there is a strong motivation for choosing an attack success rate, you can still change it back after reviewers accept your paper. ``Blend in'' your papers with those published ones will maximize your chance of getting in. }
    \label{tab:overall_res}
    \vspace{-0.5\baselineskip}
    \resizebox{0.99\textwidth}{!}{
        \begin{tabular}{lllcccccccc}
            \toprule
                                                                     &                                                              &                       & \multicolumn{6}{c}{\textbf{Attack Success Rate}}                                                                                                                                                                             \\
            \cline{4-9}
                                                                     &                                                              &                       & \multicolumn{3}{c}{\textbf{Single-Objective}}    & \multicolumn{3}{c}{\textbf{Multi-Objective}}                                                                                                                              \\
            \cline{1-3}\cline{4-6}\cline{7-9}
            \multicolumn{3}{l}{\textbf{CIFAR-10} ($\epsilon=8/255$)} & \multirow[c]{2}{*}{\makecell{APGD                                                                                                                                                                                                                                                                                   \\(1)}}             & \multirow[c]{2}{*}{\makecell{APGD\\(5)}} &   \multirow[c]{2}{*}{\makecell{ACG\\(5)}}  & \multirow[c]{2}{*}{\makecell{All\\(1)*8}}                      & \multirow[c]{2}{*}{\makecell{\textbf{MOS-8}\\\textbf{(1)}}}   & \multirow[c]{2}{*}{\makecell{\textbf{MOS-8}\\\textbf{(5)}}}                      & \multirow[c]{2}{*}{\makecell{\textbf{Diff.(5)}\\\textbf{MOS$|$CE}}}         \\
            \cline{1-3}
            \textbf{ID}                                              & \textbf{Paper}                                               & \textbf{Architecture}                                                                                                                                                                                                                                \\
            \midrule
            0                                                        & Rade~\etal (2022)~\cite{rade:2022:reducing} (\textit{ddpm})  & PreActResNet-18       & 39.17                                            & 39.28                                        & 42.45 & \cellcolor{light-gray}\textbf{42.78} (6) & 42.59 & \underline{42.77}                    & \textcolor{Green}{+3.49} \\
            1                                                        & Rade~\etal (2022)~\cite{rade:2022:reducing} (\textit{extra}) & PreActResNet-18       & 38.55                                            & 38.72                                        & 41.63 & \underline{42.21} (6)                    & 42.03 & \cellcolor{light-gray}\textbf{42.23} & \textcolor{Green}{+3.51} \\
            2                                                        & Sehwag~\etal (2022)~\cite{sehwag:2022:robust}                & ResNet-18             & 41.57                                            & 41.76                                        & 43.10 & \underline{44.16 (6)}                    & 43.79 & \cellcolor{light-gray}\textbf{44.18} & \textcolor{Green}{+2.42} \\
            \hline
            3                                                        & Chen~\etal (2020)~\cite{chen:2020:adversarial}               & ResNet-50             & 45.80                                            & 45.95                                        & 48.00 & \underline{48.04 (4)}                    & 48.09 & \cellcolor{light-gray}\textbf{48.14} & \textcolor{Green}{+3.49} \\
            \hline
            4                                                        & Gowal~\etal (2020)~\cite{gowal:2020:uncovering}              & WideResNet-28-10      & 34.31                                            & 34.46                                        & 36.39 & \cellcolor{light-gray}\textbf{36.96} (6) & 36.77 & \underline{36.95}                    & \textcolor{Green}{+2.19} \\
            5                                                        & Wang~\etal (2023)~\cite{wang:2023:better}                    & WideResNet-28-10      & 29.72                                            & 29.91                                        & 31.92 & \underline{32.44} (6)                    & 32.25 & \cellcolor{light-gray}\textbf{32.49} & \textcolor{Green}{+2.58} \\
            6                                                        & Rebuffi~\etal (2021)~\cite{rebuffi:2021:fixing}              & WideResNet-28-10      & 35.97                                            & 36.15                                        & 38.43 & \underline{39.05} (6)                    & 38.91 & \cellcolor{light-gray}\textbf{39.14} & \textcolor{Green}{+2.99} \\
            \hline
            7                                                        & Sehwag~\etal (2022)~\cite{sehwag:2022:robust}                & WideResNet-34-10      & 36.85                                            & 36.96                                        & 38.34 & \underline{39.36 (5)}                    & 38.97 & \cellcolor{light-gray}\textbf{39.38} & \textcolor{Green}{+2.43} \\
            8                                                        & Rade~\etal (2022)~\cite{rade:2022:reducing}                  & WideResNet-34-10      & 34.29                                            & 34.45                                        & 36.45 & \cellcolor{light-gray}\textbf{36.97} (6) & 36.69 & \underline{36.94}                    & \textcolor{Green}{+2.49} \\
            \hline
            9                                                        & Gowal~\etal (2021)~\cite{gowal:2021:improving}               & WideResNet-70-16      & 31.43                                            & 31.62                                        & 32.54 & \underline{33.50} (5)                    & 33.33 & \cellcolor{light-gray}\textbf{33.51} & \textcolor{Green}{+1.89} \\
            10                                                       & Gowal~\etal (2020)~\cite{gowal:2020:uncovering}              & WideResNet-70-16      & 31.89                                            & 32.07                                        & 33.34 & \cellcolor{light-gray}\textbf{33.94} (5) & 33.72 & \underline{33.92}                    & \textcolor{Green}{+1.85} \\
            11                                                       & Rebuffi~\etal (2021)~\cite{rebuffi:2021:fixing}              & WideResNet-70-16      & 30.45                                            & 30.72                                        & 32.41 & \underline{33.06} (6)                    & 32.79 & \cellcolor{light-gray}\textbf{33.10} & \textcolor{Green}{+2.38} \\
            \midrule
            \multicolumn{3}{c}{\textbf{Average Rank}}                & 5.92                                                         & 4.92                  & 4.00                                             & 1.67                                         & 3.00  & 1.33                                                                                                               \\
            \hline
            \hline
            \multicolumn{3}{l}{\textbf{ImageNet} ($\epsilon=4/255$)}                                                                                                                                                                                                                                                                                                                       \\
            \midrule
            12                                                       & Salman~\etal (2020)~\cite{salman:2020:do}                    & ResNet-18             & 70.60                                            & 70.74                                        & 72.94 & \underline{74.38} (5)                    & 74.24 & \cellcolor{light-gray}\textbf{74.52} & \textcolor{Green}{+3.87} \\
            13                                                       & Salman~\etal (2020)~\cite{salman:2020:do}                    & ResNet-50             & 61.38                                            & 61.58                                        & 62.74 & \underline{64.92} (7)                    & 64.5  & \cellcolor{light-gray}\textbf{64.94} & \textcolor{Green}{+3.36} \\
            14                                                       & Wong~\etal (2020)~\cite{wong:2020:fast}                      & ResNet-50             & 70.28                                            & 70.46                                        & 71.68 & \cellcolor{light-gray}\textbf{73.20} (5) & 72.96 & \underline{73.10}                    & \textcolor{Green}{+2.64} \\
            15                                                       & Engstrom~\etal (2019)~\cite{engstrom:2019:robustness}        & ResNet-50             & 67.62                                            & 67.82                                        & 67.72 & \cellcolor{light-gray}\textbf{70.12} (5) & 69.86 & \underline{69.92}                    & \textcolor{Green}{+2.10} \\
            16                                                       & Salman~\etal (2020)~\cite{salman:2020:do}                    & WideResNet-50-2       & 59.02                                            & 59.12                                        & 58.92 & \cellcolor{light-gray}\textbf{61.26} (5) & 60.76 & \underline{61.14}                    & \textcolor{Green}{+2.02} \\
            \midrule
            \multicolumn{3}{c}{\textbf{Average Rank}}                & 6.00                                                         & 5.00                  & 4.00                                             & 1.40                                         & 3.00  & 1.60                                                                                                               \\
            \bottomrule
        \end{tabular}
    }
\end{table*}

\lheading{Smooth Relaxation.}
Since the above problem is an NP-hard combinatorial optimization problem, we relax it by introducing a smooth relaxation.
Specifically, we relax the first objective by incorporating smooth operators in Equation~(\ref{eq:smooth_ops}) and the second objective by replacing the $\ell_0$ norm with the $\ell_1$ norm.
The relaxed problem is then:

\begin{equation} \label{eq:determine_smooth}
    \begin{aligned}
         & \min_{\bm{\beta}} \sum_{i=1}^m \mu\log(\frac{\sum_{k=1}^K e^{f_i(\delta_k)/\mu}}{\sum_{k=1}^K e^{\beta_kf_i(\delta_k)/\mu}}) + \lambda \|\bm{\beta}\|_1, \\
         & \text{s.t.} \quad \bm{\beta} \in [0,1]^K,
    \end{aligned}
\end{equation}
where $\lambda$ controls the sparsity.

The above problem is smooth and fully differentiable, and we can solve it using gradient-based methods.
After the above problem is solved, we can get the dominant example index $\bm{\beta}$. Here, we set a threhold $T$ to further make $\bm{\beta}$ binary.

\bheading{Determining Loss Synergistic Patterns.}
For every dominant perturbation $\bm{\delta}^*$, we check its contribution to the loss functions.
In particular, for each perturbation $\bm{\delta}^*$, if its $i$-th loss value $\bar{f}_i(\bm{\delta}^*) > C * \max_{\bm{\delta\in\Delta}}\bar{f}_i(\bm{\delta})$, we consider it as a contribution to the $i$-th loss function.
Thus, for every dominant perturbation, we can get a contribution combination, which we call a loss synergistic pattern.
We can record the loss synergistic pattern for each dominant perturbation across the dataset to facilitate the analysis of coupling effects between loss functions.

\subsection{Implementation: Loss Functions}
The final step of implementing our attack is to specify multiple surrogate loss functions.
We incorporate a selection of significant loss functions that are well-documented in existing literature~\cite{madry:2018:towards,croce:2020:minimally,wang:2020:improving},
along with innovative loss functions that have been identified through rigorous exploration in the domain of loss search~\cite{xia:2024:tightening,li:2022:autoloss}.
Details of these loss functions can be found in \tabref{tab:loss_funcs} and \tabref{tab:loss_notations}.

\subsection{Runtime/Efficiency Comparison}
\newcommand{\bmL}{\bm{L}}
\newcommand{\bmX}{\bm{X}}

Theoretically, MOS-Attack with $N$ loss functions and $N$ samples incurs only a constant factor more in computational expense compared to conducting $N$ single-objective attacks. Thus, the MOS-8(5) Attack may potentially offer greater efficiency than ALL-8(8) (\textbf{5} v.s. \textbf{8} examples).

\begin{table}[h]
    \centering
    \caption{Complexity analysis of gradient computation of single-objective and set-based optimization.}
    \vspace{-0.5\baselineskip}
    \resizebox{0.42\textwidth}{!}{
        \begin{tabular}{lll}
            \toprule
            Method           & Gradient Computation                                                                                             & Complexity \\
            \midrule
            Single-objective & $\frac{\partial L_i}{\partial \vx_i} = \underbrace{\frac{\partial L_i}{\partial \vh_i}}_{\substack{\text{Vector}              \\ \in \mathbb{R}^m}} \cdot \underbrace{\frac{\partial \vh_i}{\partial \vx_i}}_{\substack{\text{Jacobian} \\ \in \mathbb{R}^{m \times d}}}, \forall i$ & $\mathcal{O}\left(N \cdot \text{Cost}\left(\frac{\partial \vh}{\partial \vx}\right)\right)$\\
            \hline
            Set-based        & $\frac{\partial g}{\partial \bmX}       = \underbrace{\frac{\partial g}{\partial \bmL}}_{\substack{\text{Vector}              \\ \in \mathbb{R}^{1 \times N}}} \cdot\underbrace{\frac{\partial \bmL}{\partial \vh}}_{\substack{\text{Matrix}                                                                         \\ \in \mathbb{R}^{N \times m}}}\cdot\underbrace{\frac{\partial \vh}{\partial \bmX}}_{\substack{\text{Jacobian}                                                                       \\ \in \mathbb{R}^{m \times d \times N}}}$ & $\underbrace{\mathcal{O}\left(N \cdot \text{Cost}\left(\frac{\partial \vh}{\partial \vx}\right)\right)}_{\text{Gradient}} + \underbrace{\mathcal{O}\left(Nm\right)}_{\text{Matrix-Vector Multiplication}}$\\
            \bottomrule
            \multicolumn{3}{l}{The gradient computation cost of $\frac{\partial g}{\partial \bmL}$ and $\frac{\partial \bmL}{\partial \vh}$ is negligible compared to $\frac{\partial \vh}{\partial \vx}$.}
        \end{tabular}
    }
    \vspace{-1\baselineskip}
\end{table}

\section{Experiment}
\subsection{Experiment Setup}
\begin{figure*}
    \centering
    \begin{subfigure}{0.47\linewidth}
        \includegraphics[width=0.98\textwidth]{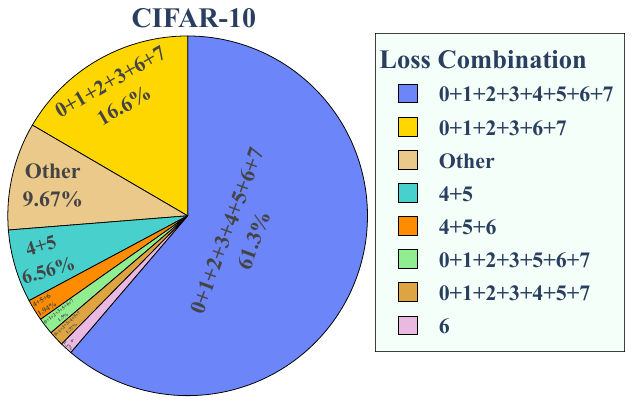}
        \caption{CIFAR-10}
    \end{subfigure}
    \hfill
    \begin{subfigure}{0.47\linewidth}
        \includegraphics[width=0.98\textwidth]{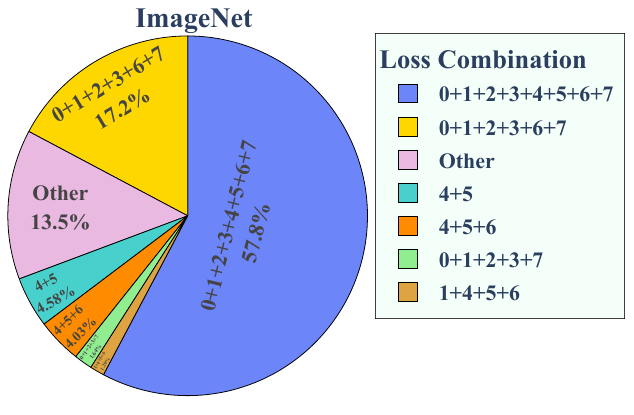}
        \caption{ImageNet}
    \end{subfigure}
    \vspace{-0.5\baselineskip}
    \caption{Occurrences of different loss synergistic patterns across CIFAR-10 and ImageNet datasets. We only retain the top patterns that account for more than $1\%$ of the adversarial examples.\label{fig:contribution_pie}}
    \vspace{-0.75\baselineskip}
\end{figure*}

\bheading{Datasets and Models.}
We employed 17 distinct from RobustBench~\cite{croce:2021:robustbench}, which includes 12 models~\cite{wang:2023:better,rebuffi:2021:fixing,sehwag:2022:robust,rade:2022:reducing,augustin:2020:adversarial,huang:2021:exploring} trained on the CIFAR-10~\cite{CIFAR10} dataset and 5 models~\cite{wong:2020:fast,salman:2020:do} based on ImageNet~\cite{deng:2009:imagenet} dataset.
For performance evaluation, we used all 10,000 test images from the CIFAR-10 validation dataset and 5,000 images from ImageNet validation dataset.
To enable direct comparison with the reported accuracy of the ACG attack, we preserved the same image indexing for the ImageNet dataset as \cite{yamamura:2022:diversified}.

\bheading{Comparative Attacks.}
For comparative purposes, we incorporate the widely recognized APGD-CE attack, the state-of-the-art ACG-CW, and the comprehensive APGD-All attack.
The latter aggregates optimal outcomes from an ensemble of eight distinct APGD attacks, each employing unique loss functions from \tabref{tab:loss_funcs}.

\bheading{Attack Parameters.}
Notably, the number of iterations for our implemented attacks, including MOS Attack, APGD-CE, and APGD-All, are uniformly set to 50.
This choice ensures thorough and rigorous testing of all methods.
Additionally, the remaining attack parameters follow the same configuration as outlined in APGD~\cite{croce:2020:reliable}.

\subsection{Overall Results}
This section presents the comparative results of our proposed MOS-8 attack alongside other competing algorithms, delineating them in terms of Attack Success Rate (ASR).
Detailed outcomes are provided in Table \tabref{tab:overall_res}.

\bheading{Single-objective \textit{v.s.} Multi-objective.}
The results demonstrate that multi-objective approaches outperform single-objective approaches. The most effective single-objective approach is the ACG-CW attack, utilizing 5 restarts and 100 attack steps; however, despite a considerably higher number of attack steps $N_{\text{iter}}=100$, it only achieved the best ASR in 3 out of 17 instances, with a rate of 3 out of 12 for CIFAR-10 and failing to succeed in any of the 5 cases for ImageNet.

\bheading{MOS-8 \textit{v.s.} APGD-All.}
The MOS-8 Attack demonstrates a slight superiority over APGD-All.
Notably, the MOS-8 Attack achieved comparable or better results with only five adversarial examples, whereas APGD-All utilized eight.
MOS-8 Attack achieved an average rank of 1.58 on CIFAR-10 and 1.60 on ImageNet, while APGD-All attained an average rank of 2.00 on CIFAR-10 and 1.40 on ImageNet.

\bheading{Loss Functions.}
APGD-All's findings underscored the superiority of loss 4-7 in \tabref{tab:loss_funcs}, as attacks using them consistently achieved the highest ASR out of 8 attack across all models on both CIFAR-10 and ImageNet.
This observation reveals the importance of selecting appropriate loss functions for adversarial attacks.

\begin{table}[t]
    \centering
    \caption{A marked discrepancy from the theoretical upper bound of set-based optimization, as estimated by comprehensive attacks.\label{tab:performance_gap}}
    \vspace{-0.75\baselineskip}
    \resizebox{0.43\textwidth}{!}{
        \begin{tabular}{l|l|cc|c|c}
            \toprule
            \textbf{ID} & \textbf{Architecture} & \makecell{\textbf{MOS-8}                               \\\textbf{(1)}} & \makecell{\textbf{MOS-8}                                                 \\\textbf{(8)}} & \makecell{\textbf{Upper}                                                 \\\textbf{Bound}} & \textbf{Diff.}                           \\
            \midrule
            0           & R-18                  & 42.59                    & 42.84 & 42.92 & -0.33/-0.08 \\
            1           & R-18                  & 42.03                    & 42.21 & 42.37 & -0.34/-0.16 \\
            2           & R-18                  & 43.79                    & 44.18 & 44.40 & -0.61/-0.22 \\
            \hline
            3           & R-50                  & 48.09                    & 48.22 & 48.36 & -0.27/-0.14 \\
            \hline
            4           & WR-28-10              & 36.77                    & 36.96 & 37.17 & -0.40/-0.21 \\
            5           & WR-28-10              & 32.25                    & 32.47 & 32.67 & -0.42/-0.30 \\
            6           & WR-28-10              & 38.91                    & 39.12 & 39.26 & -0.35/-0.14 \\
            \hline
            7           & WR-34-10              & 38.97                    & 39.39 & 39.73 & -0.76/-0.34 \\
            8           & WR-34-10              & 36.69                    & 36.95 & 37.16 & -0.47/-0.21 \\
            \hline
            9           & WR-70-16              & 33.33                    & 33.52 & 33.82 & -0.49/0.30  \\
            10          & WR-70-16              & 33.72                    & 33.95 & 34.12 & -0.40/-0.17 \\
            11          & WR-70-16              & 32.79                    & 33.08 & 33.32 & -0.53/-0.24 \\
            \bottomrule
        \end{tabular}
    }
\end{table}
% \bheading{Superiority over APGD-CE.}
% Since APGD-CE represents a widely accepted baseline for single-objective adversarial attack, the superiority of MOS-8 Attack suggest a reformulation of the standard under multi-objective setting may be necessary.

\bheading{Model Robutness.}
As the complexity of the model escalates, mirrored by the sophistication of the architecture, the performance disparity between MOS-8 Attack and APGD-CE narrows.
This indicates an incremental trend of model robustness, making them more challenging to be attacked.

\subsection{MOS Attack Upper Bound}
To evaluate the gap between the performance of our adversarial examples and the hypothetical optimal set delineated in Section~\ref{subsec:smoothset},
we conducted an array of APGD attacks on CIFAR-10 dataset.
Specifically, we implemented 8 separate APGD attacks, each employing a unique loss function and accompanied by five restarts.
For each image in the dataset,
we identified the single most effective adversarial example out of the 40 (8 attacks x 5 restarts) created.
The ASR was then calculated based on these examples to serve as an indicator of the maximum achievable performance.

\bheading{Results.}
The comparison between the MOS-8 Attack with $K=1$, $K=8$, and the upper bound is presented in \tabref{tab:performance_gap}.
Generally, the discrepancy is minimal.
Even when a single adversarial example is tailored to address all loss functions in MOS-8 Attack, near-optimal outcomes are achieved.
Additionally, leveraging eight adversarial examples brings the results within a negligible difference from the upper bound, with less than a 0.35\% gap in ASR.

% \begin{table}[t]
%     \centering
%     \caption{Extreme Performance Gap}
%     \begin{tabular}{cccccc}
%         \toprule
%         ID & MOS-1 & MOS-8 & BOA-8 & BOA-8(5) & Diff                           \\
%         \midrule
%         0  & 42.59 & 42.84 & 42.87 & 42.92    & \small{\textcolor{Red}{-0.33}} \\
%         1  & 42.03 & 42.21 & 42.35 & 42.37    & \small{\textcolor{Red}{-0.34}} \\
%         2  & 43.79 & 44.18 & 44.35 & 44.40    & \small{\textcolor{Red}{-0.61}} \\
%         3  & 48.09 & 48.22 & 48.35 & 48.36    & \small{\textcolor{Red}{-0.27}} \\
%         4  & 36.77 & 36.96 & 37.09 & 37.17    & \small{\textcolor{Red}{-0.40}} \\
%         5  & 32.25 & 32.47 & 32.63 & 32.67    & \small{\textcolor{Red}{-0.42}} \\
%         6  & 38.91 & 39.12 & 39.21 & 39.26    & \small{\textcolor{Red}{-0.35}} \\
%         7  & 38.97 & 39.39 & 39.62 & 39.73    & \small{\textcolor{Red}{-0.76}} \\
%         8  & 36.69 & 36.95 & 37.11 & 37.16    & \small{\textcolor{Red}{-0.47}} \\
%         9  & 33.33 & 33.52 & 33.80 & 33.82    & \small{\textcolor{Red}{-0.49}} \\
%         10 & 33.72 & 33.95 & 34.10 & 34.12    & \small{\textcolor{Red}{-0.40}} \\
%         11 & 32.79 & 33.08 & 33.29 & 33.32    & \small{\textcolor{Red}{-0.53}} \\
%         \bottomrule
%     \end{tabular}
% \end{table}

\begin{figure}[t]
    \begin{subfigure}{0.43\textwidth}
        \includegraphics[width=\textwidth]{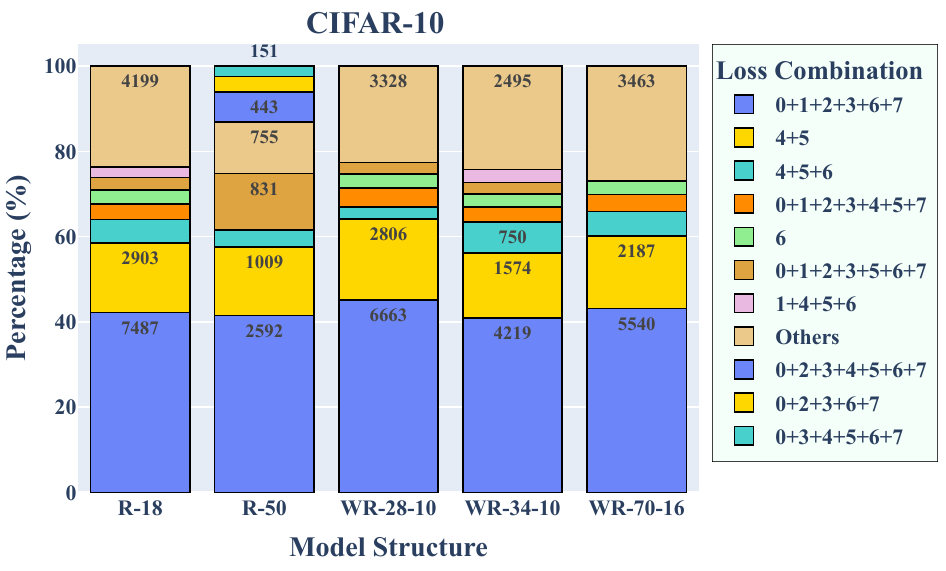}
        \caption{CIFAR-10}
    \end{subfigure}
    \begin{subfigure}{0.43\textwidth}
        \includegraphics[width=\textwidth]{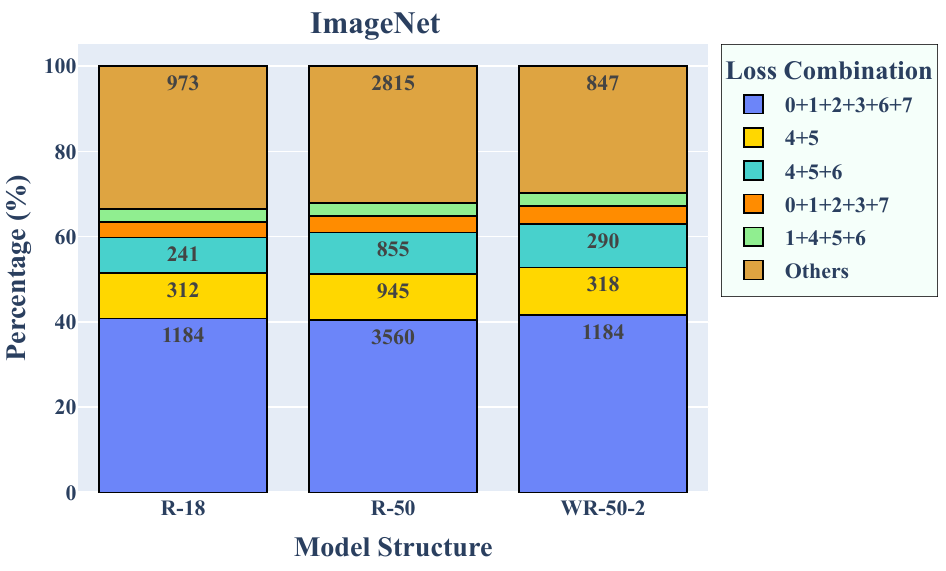}
        \caption{ImageNet}
    \end{subfigure}
    \vspace{-0.75\baselineskip}
    \caption{Detailed distribution of loss synergistic patterns across different model architectures. We only retain the top patterns that account for more than $1\%$ of the adversarial examples.\label{fig:contribution_bar}}
    \vspace{-0.75\baselineskip}
\end{figure}

\subsection{MOS Attack Analysis.}
In this section, we employ our framework to conduct an automated analysis of the relationships among various loss functions.
The solutions used for analysis is obtained from MOS-8 Attack with $K=8$ for both CIFAR-10 and ImageNet datasets.
The parameters selected were a sparsity coefficient of $\lambda=1$, a binary threshold of $T=0.85$, and a contribution threshold of $C=0.75$.

We start by identifying the synergistic patterns among loss functions for all model architectures within each dataset.
Subsequently, informed by these patterns, we design the MOS-3$^*$ attack, utilizing three selected surrogate loss functions.

\subsubsection{Loss Synergistic Pattern}
\Figref{fig:contribution_pie} depicts the synergistic loss patterns for CIFAR-10 and ImageNet.
A significant portion of the adversarial examples—$61.3\%$ for CIFAR-10 and $57.8\%$ for ImageNet—contribute to all loss functions, indicating that the majority of solutions optimize them concurrently. This observation suggests a low level of conflict among the loss functions and helps explain why employing a single loss function ($K=1$) can yield near-optimal results.

\bheading{Transferability of Synergistic Patterns.}
We extended our analysis to the transferability of these patterns across different model architectures.
We removed the common pattern containing all the losses and plotted the pattern distributions for each model architecture.
As depicted in \Figref{fig:contribution_bar}, the patterns demonstrate stability across datasets and models, with a minor exception observed in ResNet-50's patterns for the CIFAR-10 dataset, which exhibited some unique, less common patterns.

\subsubsection{MOS-3$^*$ Attack}
The predominant patterns are \textbf{0+1+2+3+6+7} and \textbf{4+5}, as they ranked first and second in both datasets, as shown in \Figref{fig:contribution_bar}.
We subsequently constructed a compact version of MOS Attack, termed MOS-3$^*$ Attack, using losses 5, 6, and 7.
For validation of the effectiveness of MOS-3$^*$ Attack, we compared it against MOS-3 Attack, which is constructed utilizing the first three loss functions.

\bheading{Results.}
As illustrated in Table \ref{tab:lite_atk}, MOS-3$^*$ Attack outperforms MOS-3 Attack.
MOS-3$^*$ Attack has achieved better performance across all models with $K=1$ adversarial example, surpassing that of MOS-3 Attack with $K=3$ adversarial examples.
Moreover, MOS-3$^*$ Attack's performance is comparable to that of MOS-8 Attack.
The above outcomes confirm the value of leveraging loss synergistic patterns to design more efficient yet effective attacks.

\begin{table}
    \centering
    \caption{The comparative results of MOS-3$^*$ Attack and MOS-3 Attack, with reference results from MOS-8 Attack.\label{tab:lite_atk}}
    \vspace{-0.75\baselineskip}
    \resizebox{0.43\textwidth}{!}{
        \begin{tabular}{l|c|c|cc|cc}
            \toprule
            \multicolumn{7}{l}{\textbf{CIFAR-10} ($\epsilon=8/255$)}                                                                                                                 \\
            \hline
            \textbf{ID} & \makecell{All                                                                                                                                              \\(1)*8}     &    \makecell{\textbf{MOS-8}\\\textbf{(5)}}         & \makecell{\textbf{MOS-3}\\\textbf{(1)}}                  & \makecell{\textbf{MOS-3}\\\textbf{(3)}}& \makecell{\textbf{MOS-3$^*$}\\\textbf{(1)}}                  & \makecell{\textbf{MOS-3$^*$}\\\textbf{(3)}}\\
            \midrule
            9           & 33.50 (5)                                & \underline{33.51}                    & 31.19 & 31.47 & \underline{33.51} & \cellcolor{light-gray}\textbf{33.60} \\
            10          & \cellcolor{light-gray}\textbf{33.94} (5) & 33.92                                & 31.63 & 31.83 & 33.91             & \underline{33.93}                    \\
            11          & 33.06 (6)                                & \cellcolor{light-gray}\textbf{33.10} & 30.23 & 30.43 & 33.03             & \underline{33.07}                    \\
            \hline
            \hline
            \multicolumn{7}{l}{\textbf{ImageNet} ($\epsilon=4/255$)}                                                                                                                 \\
            \midrule
            16          & \cellcolor{light-gray}\textbf{61.26} (5) & \underline{61.14}                    & 58.82 & 59.24 & 60.86             & 61.08                                \\
            \bottomrule
        \end{tabular}
    }
\end{table}

% \begin{table}
%     \centering
%     \begin{tabular}{ccc}
%         \toprule
%         \textbf{Datasets} & \textbf{One for all} & \textbf{Others} \\
%         CIFAR-10          &                                        \\
%         ImageNet          &
%     \end{tabular}
% \end{table}

\section{Conclusion}
Our work has introduced the MOS Attack, a novel multi-objective adversarial attack framework that effectively combines multiple surrogate loss functions to generate adversarial examples. The MOS-8 Attack, utilizing eight such functions, has shown superior performance on CIFAR-10 and ImageNet datasets compared to existing state-of-the-art methods. The framework's automated method for identifying synergistic patterns among loss functions has led to the development of the efficient MOS-3$^*$ tri-objective attack. Our contributions offer a scalable and extensible approach to adversarial machine learning, highlighting the potential for more resource-efficient and potent adversarial attack strategies in the future.

\section*{Acknowledgements}
The work described in this paper was  supported by the Research Grants Council of the Hong Kong Special Administrative Region, China [GRF Project No. CityU11215622, CityU11212524], and by Key Basic Research Foundation of Shenzhen, China (JCYJ20220818100005011).

    {
        \small
        \bibliographystyle{ieeenat_fullname}
        \bibliography{main}

@String(CVPR= {IEEE Conf. Comput. Vis. Pattern Recog.})

@String(ECCV= {Eur. Conf. Comput. Vis.})

@String(ICLR = {Int. Conf. Learn. Represent.})

@String(CVPR  = {CVPR})

@String(ECCV  = {ECCV})

@String(ICLR  = {ICLR})

@article{arora:1997:hardness,
	author  = {Sanjeev Arora and
	           L{\'{a}}szl{\'{o}} Babai and
	           Jacques Stern and
	           Z. Sweedyk},
	journal = {J. Comput. Syst. Sci.},
	title   = {The Hardness of Approximate Optima in Lattices, Codes, and Systems
	           of Linear Equations},
	year    = {1997}
}

@article{gowal:2019:alternative,
	author     = {Sven Gowal and
	              Jonathan Uesato and
	              Chongli Qin and
	              Po{-}Sen Huang and
	              Timothy A. Mann and
	              Pushmeet Kohli},
	eprinttype = {arXiv},
	journal    = {CoRR},
	title      = {An Alternative Surrogate Loss for PGD-based Adversarial Testing},
	year       = {2019}
}

@article{guo:2024:exploring,
	author  = {Guo, Ping and Gong, Cheng and Lin, Xi and Yang, Zhiyuan and Zhang, Qingfu},
	journal = {arXiv preprint arXiv:2403.05100},
	title   = {Exploring the Adversarial Frontier: Quantifying Robustness via Adversarial Hypervolume},
	year    = {2024}
}

@inproceedings{cao:2019:adversarial,
	author    = {Yulong Cao and
	             Chaowei Xiao and
	             Benjamin Cyr and
	             Yimeng Zhou and
	             Won Park and
	             Sara Rampazzi and
	             Qi Alfred Chen and
	             Kevin Fu and
	             Z. Morley Mao},
	booktitle = {Proceedings of the 2019 {ACM} {SIGSAC} Conference on Computer and
	             Communications Security, ({CCS})},
	publisher = {{ACM}},
	title     = {Adversarial Sensor Attack on LiDAR-based Perception in Autonomous
	             Driving},
	year      = {2019}
}

@inproceedings{carlini:2017:towards,
	author    = {Nicholas Carlini and
	             David A. Wagner},
	booktitle = {2017 {IEEE} Symposium on Security and Privacy, {SP}},
	publisher = {{IEEE} Computer Society},
	title     = {Towards Evaluating the Robustness of Neural Networks},
	year      = {2017}
}

@inproceedings{croce:2020:minimally,
	author    = {Francesco Croce and
	             Matthias Hein},
	booktitle = {Proceedings of the 37th International Conference on Machine Learning, {ICML}},
	title     = {Minimally distorted Adversarial Examples with a Fast Adaptive Boundary
	             Attack},
	year      = {2020}
}

@inproceedings{croce:2020:reliable,
	author    = {Francesco Croce and
	             Matthias Hein},
	booktitle = {Proceedings of the 37th International Conference on Machine Learning, {ICML}},
	publisher = {{PMLR}},
	title     = {Reliable evaluation of adversarial robustness with an ensemble of
	             diverse parameter-free attacks},
	year      = {2020}
}

@inproceedings{croce:2021:robustbench,
	author    = {Francesco Croce and
	             Maksym Andriushchenko and
	             Vikash Sehwag and
	             Edoardo Debenedetti and
	             Nicolas Flammarion and
	             Mung Chiang and
	             Prateek Mittal and
	             Matthias Hein},
	booktitle = {Proceedings of the Neural Information Processing Systems Track on
	             Datasets and Benchmarks 1, {NeurIPS Datasets and Benchmarks}},
	title     = {RobustBench: a standardized adversarial robustness benchmark},
	year      = {2021}
}

@inproceedings{deng:2019:efficient,
	author    = {Yinpeng Dong and
	             Hang Su and
	             Baoyuan Wu and
	             Zhifeng Li and
	             Wei Liu and
	             Tong Zhang and
	             Jun Zhu},
	booktitle = {{IEEE} Conference on Computer Vision and Pattern Recognition, ({CVPR})},
	publisher = {Computer Vision Foundation / {IEEE}},
	title     = {Efficient Decision-Based Black-Box Adversarial Attacks on Face Recognition},
	year      = {2019}
}

@inproceedings{goodfellow:2015:explaining,
	author    = {Ian J. Goodfellow and
	             Jonathon Shlens and
	             Christian Szegedy},
	booktitle = {3rd International Conference on Learning Representations, {ICLR}},
	title     = {Explaining and Harnessing Adversarial Examples},
	year      = {2015}
}

@inproceedings{kurakin:2017:adversarial,
	author    = {Alexey Kurakin and
	             Ian J. Goodfellow and
	             Samy Bengio},
	booktitle = {5th International Conference on Learning Representations, {ICLR}},
	title     = {Adversarial examples in the physical world},
	year      = {2017}
}

@inproceedings{madry:2018:towards,
	author    = {Aleksander Madry and
	             Aleksandar Makelov and
	             Ludwig Schmidt and
	             Dimitris Tsipras and
	             Adrian Vladu},
	booktitle = {6th International Conference on Learning Representations, {ICLR}},
	title     = {Towards Deep Learning Models Resistant to Adversarial Attacks},
	year      = {2018}
}

@inproceedings{sriramanan:2020:guided,
	author    = {Gaurang Sriramanan and
	             Sravanti Addepalli and
	             Arya Baburaj and
	             Venkatesh Babu R.},
	booktitle = {Advances in Neural Information Processing Systems 33: Annual Conference
	             on Neural Information Processing Systems 2020, {NeurIPS}},
	title     = {Guided Adversarial Attack for Evaluating and Enhancing Adversarial
	             Defenses},
	year      = {2020}
}

@inproceedings{szegedy:2014:intriguing,
	author    = {Christian Szegedy and
	             Wojciech Zaremba and
	             Ilya Sutskever and
	             Joan Bruna and
	             Dumitru Erhan and
	             Ian J. Goodfellow and
	             Rob Fergus},
	booktitle = {2nd International Conference on Learning Representations, {ICLR}},
	title     = {Intriguing properties of neural networks},
	year      = {2014}
}

@inproceedings{wang:2020:improving,
	author    = {Yisen Wang and
	             Difan Zou and
	             Jinfeng Yi and
	             James Bailey and
	             Xingjun Ma and
	             Quanquan Gu},
	booktitle = {8th International Conference on Learning Representations, {ICLR}},
	publisher = {OpenReview.net},
	title     = {Improving Adversarial Robustness Requires Revisiting Misclassified
	             Examples},
	year      = {2020}
}

@inproceedings{yamamura:2022:diversified,
	author    = {Keiichiro Yamamura and
	             Haruki Sato and
	             Nariaki Tateiwa and
	             Nozomi Hata and
	             Toru Mitsutake and
	             Issa Oe and
	             Hiroki Ishikura and
	             Katsuki Fujisawa},
	booktitle = {International Conference on Machine Learning, {ICML}},
	publisher = {{PMLR}},
	series    = {Proceedings of Machine Learning Research},
	title     = {Diversified Adversarial Attacks based on Conjugate Gradient Method},
	year      = {2022}
}

@inproceedings{zhang:2019:theoretically,
	author    = {Hongyang Zhang and
	             Yaodong Yu and
	             Jiantao Jiao and
	             Eric P. Xing and
	             Laurent El Ghaoui and
	             Michael I. Jordan},
	booktitle = {Proceedings of the 36th International Conference on Machine Learning,
	             {ICML}},
	publisher = {{PMLR}},
	title     = {Theoretically Principled Trade-off between Robustness and Accuracy},
	year      = {2019}
}

@inproceedings{lin:2024:smooth,
	author    = {Xi Lin and
	             Xiaoyuan Zhang and
	             Zhiyuan Yang and
	             Fei Liu and
	             Zhenkun Wang and
	             Qingfu Zhang},
	booktitle = {Forty-first International Conference on Machine Learning, {ICML}},
	title     = {Smooth Tchebycheff Scalarization for Multi-Objective Optimization},
	year      = {2024}
}

@article{lin:2024:few,
	author     = {Xi Lin and
	              Yilu Liu and
	              Xiaoyuan Zhang and
	              Fei Liu and
	              Zhenkun Wang and
	              Qingfu Zhang},
	eprinttype = {arXiv},
	journal    = {CoRR},
	title      = {Few for Many: Tchebycheff Set Scalarization for Many-Objective Optimization},
	year       = {2024}
}

@inproceedings{xia:2024:tightening,
	author    = {Pengfei Xia and
	             Ziqiang Li and
	             Bin Li},
	booktitle = {Proceedings of the Genetic and Evolutionary Computation Conference
	             Companion, {GECCO} 2024, Melbourne, VIC, Australia, July 14-18, 2024},
	editor    = {Xiaodong Li and
	             Julia Handl},
	title     = {Tightening the Approximation Error of Adversarial Risk with Auto Loss
	             Function Search},
	year      = {2024}
}

@techreport{CIFAR10,
	author      = {A. Krizhevsky},
	institution = {Univ. Toronto},
	title       = {{Learning Multiple Layers of Features from Tiny Images}},
	type        = {Technical Report},
	year        = {2009}
}

@inproceedings{wang:2023:better,
	author    = {Zekai Wang and
	             Tianyu Pang and
	             Chao Du and
	             Min Lin and
	             Weiwei Liu and
	             Shuicheng Yan},
	bibsource = {dblp computer science bibliography, https://dblp.org},
	booktitle = {International Conference on Machine Learning, ({ICML})},
	publisher = {{PMLR}},
	series    = {Proceedings of Machine Learning Research},
	title     = {Better Diffusion Models Further Improve Adversarial Training},
	year      = {2023}
}

@article{rebuffi:2021:fixing,
	author     = {Sylvestre{-}Alvise Rebuffi and
	              Sven Gowal and
	              Dan A. Calian and
	              Florian Stimberg and
	              Olivia Wiles and
	              Timothy A. Mann},
	eprinttype = {arXiv},
	journal    = {CoRR},
	title      = {Fixing Data Augmentation to Improve Adversarial Robustness},
	year       = {2021}
}

@inproceedings{sehwag:2022:robust,
	author    = {Vikash Sehwag and
	             Saeed Mahloujifar and
	             Tinashe Handina and
	             Sihui Dai and
	             Chong Xiang and
	             Mung Chiang and
	             Prateek Mittal},
	booktitle = {The Tenth International Conference on Learning Representations, ({ICLR})},
	publisher = {OpenReview.net},
	title     = {Robust Learning Meets Generative Models: Can Proxy Distributions Improve
	             Adversarial Robustness?},
	year      = {2022}
}

@inproceedings{rade:2022:reducing,
	author    = {Rahul Rade and
	             Seyed{-}Mohsen Moosavi{-}Dezfooli},
	booktitle = {The Tenth International Conference on Learning Representations, ({ICLR})},
	publisher = {OpenReview.net},
	title     = {Reducing Excessive Margin to Achieve a Better Accuracy vs. Robustness
	             Trade-off},
	year      = {2022}
}

@inproceedings{augustin:2020:adversarial,
	author    = {Maximilian Augustin and
	             Alexander Meinke and
	             Matthias Hein},
	booktitle = {Computer Vision - {ECCV} 2020 - 16th European Conference},
	publisher = {Springer},
	series    = {Lecture Notes in Computer Science},
	title     = {Adversarial Robustness on In- and Out-Distribution Improves Explainability},
	year      = {2020}
}

@inproceedings{huang:2021:exploring,
	author    = {Hanxun Huang and
	             Yisen Wang and
	             Sarah M. Erfani and
	             Quanquan Gu and
	             James Bailey and
	             Xingjun Ma},
	booktitle = {Advances in Neural Information Processing Systems 34: Annual Conference
	             on Neural Information Processing Systems 2021, ({NeurIPS})},
	title     = {Exploring Architectural Ingredients of Adversarially Robust Deep Neural
	             Networks},
	year      = {2021}
}

@inproceedings{wong:2020:fast,
	author    = {Eric Wong and
	             Leslie Rice and
	             J. Zico Kolter},
	booktitle = {8th International Conference on Learning Representations, {ICLR}},
	publisher = {OpenReview.net},
	title     = {Fast is better than free: Revisiting adversarial training},
	year      = {2020}
}

@inproceedings{salman:2020:do,
	author    = {Hadi Salman and
	             Andrew Ilyas and
	             Logan Engstrom and
	             Ashish Kapoor and
	             Aleksander Madry},
	booktitle = {Advances in Neural Information Processing Systems 33: Annual Conference
	             on Neural Information Processing Systems 2020, NeurIPS},
	title     = {Do Adversarially Robust ImageNet Models Transfer Better?},
	year      = {2020}
}

@misc{engstrom:2019:robustness,
	author = {Logan Engstrom and Andrew Ilyas and Hadi Salman and Shibani Santurkar and Dimitris Tsipras},
	title  = {Robustness (Python Library)},
	url    = {https://github.com/MadryLab/robustness},
	year   = {2019}
}

@inproceedings{chen:2020:adversarial,
	author    = {Tianlong Chen and
	             Sijia Liu and
	             Shiyu Chang and
	             Yu Cheng and
	             Lisa Amini and
	             Zhangyang Wang},
	booktitle = {2020 {IEEE/CVF} Conference on Computer Vision and Pattern Recognition,
	             {CVPR} 2020, Seattle, WA, USA, June 13-19, 2020},
	publisher = {Computer Vision Foundation / {IEEE}},
	title     = {Adversarial Robustness: From Self-Supervised Pre-Training to Fine-Tuning},
	year      = {2020}
}

@article{gowal:2020:uncovering,
	author     = {Sven Gowal and
	              Chongli Qin and
	              Jonathan Uesato and
	              Timothy A. Mann and
	              Pushmeet Kohli},
	bibsource  = {dblp computer science bibliography, https://dblp.org},
	eprint     = {2010.03593},
	eprinttype = {arXiv},
	journal    = {CoRR},
	title      = {Uncovering the Limits of Adversarial Training against Norm-Bounded
	              Adversarial Examples},
	url        = {https://arxiv.org/abs/2010.03593},
	volume     = {abs/2010.03593},
	year       = {2020}
}

@inproceedings{gowal:2021:improving,
	author    = {Sven Gowal and
	             Sylvestre{-}Alvise Rebuffi and
	             Olivia Wiles and
	             Florian Stimberg and
	             Dan Andrei Calian and
	             Timothy A. Mann},
	bibsource = {dblp computer science bibliography, https://dblp.org},
	booktitle = {Advances in Neural Information Processing Systems 34: Annual Conference
	             on Neural Information Processing Systems 2021, NeurIPS},
	title     = {Improving Robustness using Generated Data},
	year      = {2021}
}

@article{nikolaos:2022:alternating,
	author     = {Nikolaos Antoniou and
	              Efthymios Georgiou and
	              Alexandros Potamianos},
	eprinttype = {arXiv},
	journal    = {CoRR},
	title      = {Alternating Objectives Generates Stronger PGD-Based Adversarial Attacks},
	year       = {2022}
}

@book{miettinen2012nonlinear,
	author    = {Miettinen, Kaisa},
	publisher = {Springer Science \& Business Media},
	title     = {Nonlinear multiobjective optimization},
	volume    = {12},
	year      = {2012}
}

@article{beck:2012:smoothing,
	author  = {Amir Beck and
	           Marc Teboulle},
	journal = {{SIAM} J. Optim.},
	title   = {Smoothing and First Order Methods: {A} Unified Framework},
	year    = {2012}
}

@inproceedings{he:2016:deep,
	author    = {Kaiming He and
	             Xiangyu Zhang and
	             Shaoqing Ren and
	             Jian Sun},
	booktitle = {{IEEE} Conference on Computer Vision and Pattern Recognition,
	             {CVPR}},
	publisher = {{IEEE} Computer Society},
	title     = {Deep Residual Learning for Image Recognition},
	year      = {2016}
}

@inproceedings{sabour:2017:dynamic,
	author    = {Sara Sabour and
	             Nicholas Frosst and
	             Geoffrey E. Hinton},
	booktitle = {Advances in Neural Information Processing Systems 30: Annual Conference
	             on Neural Information Processing Systems},
	title     = {Dynamic Routing Between Capsules},
	year      = {2017}
}

@inproceedings{deng:2009:imagenet,
	author    = {Jia Deng and
	             Wei Dong and
	             Richard Socher and
	             Li{-}Jia Li and
	             Kai Li and
	             Li Fei{-}Fei},
	booktitle = {2009 {IEEE} Computer Society Conference on Computer Vision and Pattern
	             Recognition {CVPR}},
	publisher = {{IEEE} Computer Society},
	title     = {ImageNet: {A} large-scale hierarchical image database},
	year      = {2009}
}

@article{laleh:2022:adversarial,
	author    = {Ghaffari Laleh, Narmin and Truhn, Daniel and Veldhuizen, Gregory Patrick and Han, Tianyu and van Treeck, Marko and Buelow, Roman D and Langer, Rupert and Dislich, Bastian and Boor, Peter and Schulz, Volkmar and others},
	journal   = {Nature communications},
	publisher = {Nature Publishing Group UK London},
	title     = {Adversarial attacks and adversarial robustness in computational pathology},
	year      = {2022}
}

@inproceedings{li:2022:autoloss,
	author    = {Hao Li and
	             Tianwen Fu and
	             Jifeng Dai and
	             Hongsheng Li and
	             Gao Huang and
	             Xizhou Zhu},
	booktitle = {{IEEE/CVF} Conference on Computer Vision and Pattern Recognition,
	             {CVPR}},
	publisher = {{IEEE}},
	title     = {AutoLoss-Zero: Searching Loss Functions from Scratch for Generic Tasks},
	year      = {2022}
}

@inproceedings{williams:2023:black-box,
	author    = {Phoenix Neale Williams and
	             Ke Li},
	booktitle = {{IEEE/CVF} Conference on Computer Vision and Pattern Recognition,
	             {CVPR}},
	publisher = {{IEEE}},
	title     = {Black-Box Sparse Adversarial Attack via Multi-Objective Optimisation
	             {CVPR} Proceedings},
	year      = {2023}
}

@article{liu:2024:effective,
	author  = {Shengcai Liu and
	           Ning Lu and
	           Wenjing Hong and
	           Chao Qian and
	           Ke Tang},
	journal = {{ACM} Trans. Evol. Learn. Optim.},
	title   = {Effective and Imperceptible Adversarial Textual Attack Via Multi-objectivization},
	year    = {2024}
}

@article{zhang:2007:moead,
	author  = {Qingfu Zhang and
	           Hui Li},
	journal = {{IEEE} Trans. Evol. Comput.},
	title   = {{MOEA/D:} {A} Multiobjective Evolutionary Algorithm Based on Decomposition},
	year    = {2007}
}

@inproceedings{guo:2024:lautoda,
	author    = {Ping Guo and
	             Fei Liu and
	             Xi Lin and
	             Qingchuan Zhao and
	             Qingfu Zhang},
	booktitle = {Proceedings of the Genetic and Evolutionary Computation Conference
	             Companion, {GECCO}},
	publisher = {{ACM}},
	title     = {L-AutoDA: Large Language Models for Automatically Evolving Decision-based
	             Adversarial Attacks},
	year      = {2024}
}

@inproceedings{fu:2022:autoda,
	author    = {Qi{-}An Fu and
	             Yinpeng Dong and
	             Hang Su and
	             Jun Zhu and
	             Chao Zhang},
	booktitle = {31st {USENIX} Security Symposium, {USENIX} Security},
	publisher = {{USENIX} Association},
	title     = {AutoDA: Automated Decision-based Iterative Adversarial Attacks},
	year      = {2022}
}

@article{wang:2024:amulti,
	author     = {Hanrui Wang and
	              Shuo Wang and
	              Cunjian Chen and
	              Massimo Tistarelli and
	              Zhe Jin},
	eprint     = {2408.08205},
	eprinttype = {arXiv},
	journal    = {CoRR},
	title      = {A Multi-task Adversarial Attack Against Face Authentication},
	year       = {2024}
}

@inproceedings{deist:2023:multi,
	author    = {Timo M. Deist and
	             Monika Grewal and
	             Frank J. W. M. Dankers and
	             Tanja Alderliesten and
	             Peter A. N. Bosman},
	booktitle = {Evolutionary Multi-Criterion Optimization - 12th International Conference, {EMO}},
	publisher = {Springer},
	title     = {Multi-objective Learning Using {HV} Maximization},
	year      = {2023}
}

@inproceedings{Suzuki:2019:adversarial,
	author    = {Takahiro Suzuki and
	             Shingo Takeshita and
	             Satoshi Ono},
	booktitle = {{IEEE} Congress on Evolutionary Computation, {CEC}},
	publisher = {{IEEE}},
	title     = {Adversarial Example Generation using Evolutionary Multi-objective
	             Optimization},
	year      = {2019}
}

@inproceedings{baia:2021:effective,
	author    = {Alina Elena Baia and
	             Gabriele Di Bari and
	             Valentina Poggioni},
	booktitle = {Applications of Evolutionary Computation - 24th International Conference,
	             EvoApplications},
	publisher = {Springer},
	series    = {Lecture Notes in Computer Science},
	title     = {Effective Universal Unrestricted Adversarial Attacks Using a {MOE}
	             Approach},
	year      = {2021}
}

@inproceedings{antoniou:2020:square,
	author    = {Maksym Andriushchenko and
	             Francesco Croce and
	             Nicolas Flammarion and
	             Matthias Hein},
	booktitle = {Computer Vision - {ECCV} 2020 - 16th European Conference},
	publisher = {Springer},
	series    = {Lecture Notes in Computer Science},
	title     = {Square Attack: {A} Query-Efficient Black-Box Adversarial Attack via
	             Random Search},
	year      = {2020}
}

@inproceedings{wang:2015:pareto,
	author    = {Rui Wang and
	             Qingfu Zhang and
	             Tao Zhang},
	booktitle = {Evolutionary Multi-Criterion Optimization - 8th International Conference,
	             {EMO}},
	publisher = {Springer},
	series    = {Lecture Notes in Computer Science},
	title     = {Pareto Adaptive Scalarising Functions for Decomposition Based Algorithms},
	year      = {2015}
}

@article{deb:2002:fast,
	author    = {Kalyanmoy Deb and
	             Samir Agrawal and
	             Amrit Pratap and
	             T. Meyarivan},
	journal   = {{IEEE} Trans. Evol. Comput.},
	title     = {A fast and elitist multiobjective genetic algorithm: {NSGA-II}},
	year      = {2002}
}
    }


\end{document}